\documentclass{article}

\usepackage{microtype}
\usepackage{graphicx}
\usepackage{subcaption}
\usepackage{booktabs}
\usepackage{multirow}
\usepackage{placeins}
\usepackage{amsmath}
\usepackage{amssymb}
\usepackage{mathtools}
\usepackage{nicefrac}
\usepackage{xcolor}
\usepackage{hyperref}
\usepackage{lmodern}
\usepackage{tikz}
\usetikzlibrary{positioning,calc,fit,backgrounds,arrows.meta,shapes.geometric,shapes.misc}

\definecolor{colGenomics}{HTML}{B7CEEB}
\definecolor{colProteomics}{HTML}{F5C9A0}
\definecolor{colMetabolomics}{HTML}{D4C5EC}
\definecolor{colExperiment}{HTML}{F5D547}
\definecolor{colOntology}{HTML}{D5D5D5}
\definecolor{colInk}{HTML}{1F2937}
\definecolor{colEdge}{HTML}{374151}

\usepackage[accepted]{icml2026}

\usepackage[capitalize,noabbrev]{cleveref}

\newcommand{\canopy}{\textsc{Canopy}}

\icmltitlerunning{Canopy: A Heterograph Foundation Model for Metabolic Engineering}

\begin{document}

\twocolumn[
  \icmltitle{Canopy: A Heterograph Foundation Model for Metabolic Engineering}

  \icmlsetsymbol{equal}{*}

  \begin{icmlauthorlist}
    \icmlauthor{Jake Bowden}{twig}
    \icmlauthor{Laurence Legon}{twig}
    \icmlauthor{Satnam Surae}{twig}
  \end{icmlauthorlist}

  \icmlaffiliation{twig}{Twig Bio, London, United Kingdom}

  \icmlcorrespondingauthor{Jake Bowden}{jake.bowden@twig.bio}
  \icmlcorrespondingauthor{Satnam Surae}{satnam.surae@twig.bio}

  \icmlkeywords{Knowledge Graphs, Heterogeneous Graph Transformer,
    Foundation Models, Metabolic Engineering, Self-Supervised Learning}

  \vskip 0.3in
]

% icml2026.sty's running-title size check is overly strict (it measures the
% title in a \vbox, whose height includes \baselineskip ~11pt and so always
% exceeds the 6.25pt threshold). Re-assert our short running title after
% \icmltitle has clobbered it with the "Title Suppressed" placeholder.
\makeatletter
\gdef\@icmltitlerunning{Canopy: A Heterograph Foundation Model for Metabolic Engineering}
\makeatother

\printAffiliationsAndNotice{}

%% ===================================================================
%%  ABSTRACT
%% ===================================================================
\begin{abstract}
Designing microbial strains that produce high-value chemicals at
commercially viable titers remains a central challenge in metabolic
engineering. Existing computational approaches either rely on
stoichiometric constraint-based models that cannot learn from
experimental data, or apply tabular machine learning to hand-crafted
features that discard the relational structure of biological
knowledge. We present \canopy{}, a heterogeneous graph foundation
model that integrates ten public and proprietary data sources into a
unified knowledge graph (KG) of 6.9\,M nodes across 13 types and 34 edge types,
covering genes, proteins, metabolites, reactions, pathways, strains,
and fermentation experiments. Node features are encoded through
domain-specific foundation models (ESM-2 for protein sequences,
MoLFormer for chemical SMILES, and PubMedBERT for biomedical
text), yielding a multi-modal representation within a single graph.
We pretrain a Heterogeneous Graph Transformer (HGT) augmented with SignNet
positional encodings, Jumping Knowledge aggregation, and virtual
nodes using four self-supervised objectives (link
prediction, masked node modelling, distance prediction, and
contrastive experiment clustering), balanced via learned homoscedastic
uncertainty weighting. On the downstream task of fermentation titer
prediction, frozen \canopy{} embeddings achieve $R^{2} = 0.41$ with
a lightweight probe, outperforming tabular baselines (best
$R^{2} = 0.24$) and homogeneous GNN variants.
\end{abstract}

%% ===================================================================
%%  1  INTRODUCTION
%% ===================================================================
\section{Introduction}

The bioeconomy depends on engineered microorganisms that convert
renewable feedstocks into fuels, pharmaceuticals, and specialty
chemicals. The design--build--test--learn (DBTL) cycle that underpins
strain engineering is slow: a single round of genetic modification,
fermentation, and analytical characterisation can take weeks to
months \citep{doi:10.1021/acssynbio.9b00020}, and the combinatorial
space of candidate modifications grows
exponentially in the number of target genes. Computational tools that
prioritise genetic interventions before wet-lab experiments would
shorten this loop.

The dominant computational paradigm for strain design is
constraint-based modelling via genome-scale metabolic models (GEMs).
Flux balance analysis (FBA) and its extensions predict steady-state
fluxes under stoichiometric and thermodynamic constraints, enabling
\emph{in silico} knockout analysis through tools such as OptKnock
\citep{burgard_optknock_2003} and StrainDesign
\citep{schneider_straindesign_2022}. While GEMs encode mechanistic knowledge of metabolism,
they cannot incorporate experimental measurements of titer, rate, and
yield; they ignore regulatory, expression-level, and
environmental effects; and they treat each organism in isolation
without using cross-organism transfer.

Machine learning offers a complementary approach. Previous work has
applied random forests, gradient boosting, and neural networks to
predict fermentation titer from features extracted from strain
descriptions \citep{oyetunde_machine_2019,czajka_integrated_2021}. These tabular approaches
discard the relational structure that links genes to proteins,
proteins to reactions, reactions to pathways, and pathways to
production phenotypes. Graph neural networks have been applied to
metabolic networks for gene essentiality \citep{hasibi_integration_2024} and
site-of-metabolism prediction \citep{porokhin_using_2023}, but these operate on
single-organism reaction graphs rather than on cross-organism
knowledge graphs.

Meanwhile, the broader ML community has advanced heterogeneous graph
transformers \citep{DBLP:journals/corr/abs-2003-01332}, self-supervised graph pretraining
\citep{hu_strategies_2020}, and domain-specific foundation models for
proteins \citep{lin_evolutionary-scale_2023} and small molecules
\citep{ross_large-scale_2021}. Biomedical knowledge graphs such as PrimeKG
\citep{chandak_building_2023} have enabled GNN-based drug repurposing, but no
analogous resource or foundation model exists for the metabolic
engineering domain, which sits at the intersection of microbial genetics, enzyme
biochemistry, and fermentation science.

We introduce \canopy{}, a heterogeneous graph foundation model for
metabolic engineering. Our contributions are:

\begin{itemize}\itemsep2pt
  \item \textbf{A metabolic-engineering knowledge graph} integrating
    ten data sources (MetaNetX, GO, UniRef, InterPro, NCBI Taxonomy
    and Genomics, KEGG, an in-house laboratory information management
    system (LIMS) of unpublished DBTL records, literature-mined
    experiments, and transcriptomics) via BioCypher
    \citep{lobentanzer_democratising_2023} into a
    unified heterogeneous graph with 13 node types and 34 edge types.
  \item \textbf{Multi-modal feature encoding} using a schema-driven
    dispatch system that routes protein sequences to ESM-2 (650M),
    SMILES strings to MoLFormer-XL, free text to PubMedBERT, and
    numeric features to normalised scalars within a single graph.
  \item \textbf{An augmented Heterogeneous Graph Transformer}
    combining HGTConv with SignNet positional encodings, random walk
    structural encodings, per-type feed-forward networks (FFNs) with learnable residual
    scaling, Jumping Knowledge aggregation, and virtual nodes.
  \item \textbf{Four-objective self-supervised pretraining} with
    learned uncertainty weighting: link prediction, masked node
    modelling, distance prediction, and a contrastive experiment-pair
    loss.
  \item \textbf{Downstream evaluation} on fermentation titer
    prediction, showing that learned graph representations
    outperform tabular baselines and provide the conditioning
    substrate for downstream generative strain-design pipelines.
\end{itemize}

%% ===================================================================
%%  2  RELATED WORK
%% ===================================================================
\section{Related Work}
\begin{figure*}[!t]
  \centering
  \resizebox{\textwidth}{!}{\sffamily%
  \begin{tikzpicture}[
    every node/.style = {font=\sffamily\scriptsize, inner sep=0pt},
    ntext/.style       = {align=center, font=\sffamily\tiny\bfseries,
                          inner sep=0.5pt},
    n-base/.style      = {draw=colInk!75, line width=0.4pt, ntext},
    n-genomics/.style    = {n-base, regular polygon, regular polygon sides=6,
                            minimum size=9mm, fill=colGenomics},
    n-proteomics/.style  = {n-base, chamfered rectangle,
                            minimum width=11mm, minimum height=7mm,
                            chamfered rectangle xsep=1.2pt,
                            chamfered rectangle ysep=1.2pt,
                            fill=colProteomics},
    n-metabolomics/.style= {n-base, circle, minimum size=9mm,
                            fill=colMetabolomics},
    n-experiment/.style  = {n-base, rectangle, rounded corners=1pt,
                            minimum width=11mm, minimum height=7mm,
                            fill=colExperiment},
    n-ontology/.style    = {n-base, regular polygon, regular polygon sides=5,
                            minimum size=9mm, fill=colOntology},
    rel/.style         = {-{Latex[length=1.2mm,width=0.85mm]},
                          draw=colEdge, line width=0.35pt},
    relabel/.style     = {font=\sffamily\tiny, color=colEdge!85,
                          inner sep=0.7pt, fill=white,
                          fill opacity=0.9, text opacity=1},
    legend-base/.style = {n-base, minimum height=3.6mm, minimum width=3.6mm,
                          inner sep=1.5pt, font=\sffamily\tiny},
    legend-genomics/.style    = {legend-base, regular polygon,
                                 regular polygon sides=6, fill=colGenomics},
    legend-proteomics/.style  = {legend-base, chamfered rectangle,
                                 fill=colProteomics},
    legend-metabolomics/.style= {legend-base, circle, fill=colMetabolomics},
    legend-experiment/.style  = {legend-base, rectangle, rounded corners=1pt,
                                 fill=colExperiment},
    legend-ontology/.style    = {legend-base, regular polygon,
                                 regular polygon sides=5, fill=colOntology},
  ]
    \node[n-genomics]    (taxa)     at (0.6,  0.0) {Taxa};
    \node[n-experiment]  (tstudy)   at (4.4,  0.0) {Trscr.\\study};
    \node[n-experiment]  (tres)     at (8.2,  0.0) {Trscr.\\result};
    \node[n-genomics]    (chassis)  at (0.6, -1.5) {Chassis};
    \node[n-genomics]    (gene)     at (4.4, -1.5) {Gene};
    \node[n-proteomics]  (uniref)   at (8.2, -1.5) {UniRef\\Cluster};
    \node[n-ontology]    (go)       at (12.0,-1.5) {Gene\\Ontol.};

    \node[n-proteomics]  (interpro) at (12.0,-3.0) {InterPro\\Domain};

    \node[n-experiment]  (strain)   at (0.6, -3.0) {Strain};
    \node[n-experiment]  (pathway)  at (4.4, -3.0) {Pathway};
    \node[n-metabolomics](rxn)      at (8.2, -3.0) {Rxn};

    \node[n-experiment]  (exp)      at (0.6, -4.5) {Exp.};
    \node[n-metabolomics](met)      at (4.4, -4.5) {Metab.};

    \draw[rel] (taxa)    -- node[relabel]            {child\_of}       (chassis);
    \draw[rel] (chassis) -- node[relabel]            {used\_in}        (tstudy);
    \draw[rel] (tstudy)  -- node[relabel]            {has\_result}     (tres);
    \draw[rel] (tres)    -- node[relabel]            {measures}        (gene);
    \draw[rel] (chassis) -- node[relabel]            {contains}        (gene);
    \draw[rel] (gene)    -- node[relabel]            {encodes}         (uniref);
    \draw[rel] (uniref)  -- node[relabel]            {has\_domain}     (interpro);
    \draw[rel] (uniref)  -- node[relabel]            {described\_by}   (go);
    \draw[rel] (chassis) -- node[relabel]            {base\_for}       (strain);
    \draw[rel] (strain)  -- node[relabel]            {used\_in}        (exp);
    \draw[rel] (strain)  -- node[relabel]            {contains}        (pathway);
    \draw[rel] (pathway) -- node[relabel]            {has}             (rxn);
    \draw[rel] (uniref)  -- node[relabel]            {catalyses}       (rxn);
    \draw[rel] (met)     -- node[relabel, pos=0.55]  {participates\_in}(rxn);
    \draw[rel] (strain)  -- node[relabel, pos=0.55]  {produces}        (met);
    \draw[rel] (strain) -- node[relabel, pos=0.5] {modification} (gene);

    \draw[rel] (go) edge[loop above, looseness=7, in=70, out=110,
                         min distance=2.8mm]
                node[relabel, yshift=-3pt] {child\_of} (go);

    \draw[rel] (taxa) edge[loop above, looseness=7, in=70, out=110,
                           min distance=2.8mm]
                node[relabel, yshift=-3pt] {child\_of} (taxa);

    \begin{scope}[shift={(0.6, -5.55)}]
      \node[legend-genomics]     (lg-gen) at (0.0, 0) {};
      \node[font=\sffamily\tiny, anchor=west] at ($(lg-gen.east)+(0.05,0)$) {Genomics};
      \node[legend-proteomics]   (lg-pro) at (2.4, 0) {};
      \node[font=\sffamily\tiny, anchor=west] at ($(lg-pro.east)+(0.05,0)$) {Proteomics};
      \node[legend-metabolomics] (lg-met) at (4.8, 0) {};
      \node[font=\sffamily\tiny, anchor=west] at ($(lg-met.east)+(0.05,0)$) {Metabolomics};
      \node[legend-experiment]   (lg-exp) at (7.5, 0) {};
      \node[font=\sffamily\tiny, anchor=west] at ($(lg-exp.east)+(0.05,0)$) {Experiment};
      \node[legend-ontology]     (lg-ont) at (10.0, 0) {};
      \node[font=\sffamily\tiny, anchor=west] at ($(lg-ont.east)+(0.05,0)$) {Ontology};
    \end{scope}
  \end{tikzpicture}}
  \caption{
    \canopy{} heterograph schema. Each node type has a distinct shape
    and is coloured by data domain (genomics, proteomics, metabolomics,
    experiment, ontology); each edge is typed by relation.
    Strain-to-gene \textsf{modification} edges include knockouts,
    knockins, and knockdowns.
  }
  \label{fig:schema}
\end{figure*}
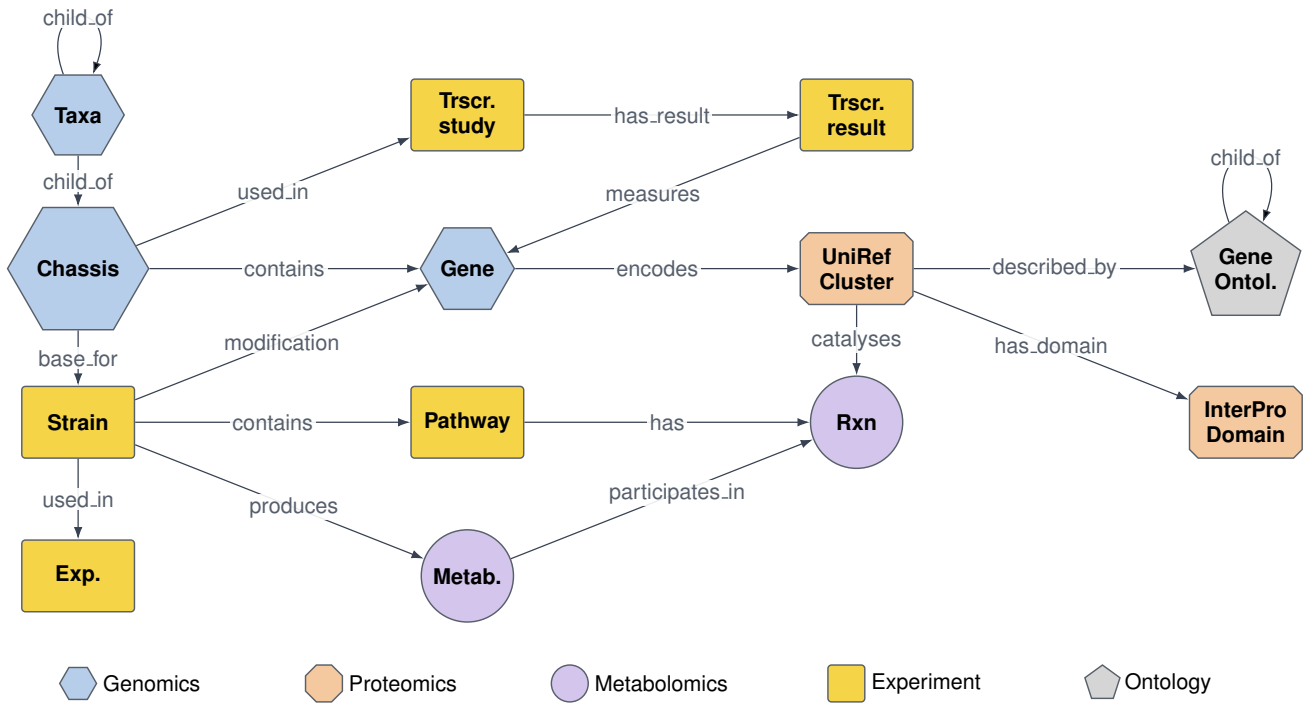

\paragraph{Foundation models across biological scales.}
Domain-specific foundation models now span genomes
\citep{doi:10.1126/science.ado9336,brixi_genome_2025}, protein sequence
\citep{lin_evolutionary-scale_2023,hayes_simulating_2025} and structure
\citep{su_saprot_2024,abramson_accurate_2024}, molecules
\citep{ross_large-scale_2021}, and single-cell transcriptomics
\citep{cui_scgpt_2024,hao_large-scale_2024}. Each captures a single
modality at a single scale; none are jointly trained over the
relational structure that links genes to enzymes to metabolites to
fermentation outcomes. \canopy{} embeds frozen ESM-2, MoLFormer, and
PubMedBERT features inside a heterogeneous KG, treating these models
as feature encoders within a graph that crosses scales.

\paragraph{Predictive and generative models for strain, pathway, and enzyme design.}
Kinetic GEMs such as k-ecoli457 \citep{khodayari_genome-scale_2016}
extend FBA with parameterised rate equations fit to fluxomic data and
predict product yields more accurately than purely stoichiometric
models, but their parameterisation is organism-specific and does not
transfer across strains.
Tabular ML on hand-crafted features remains dominant for titer
prediction \citep{oyetunde_machine_2019,czajka_integrated_2021,radivojevic_machine_2020}; graph
methods so far have been single-organism
\citep{hasibi_integration_2024,xin_gene-metabolite_2024} or relied on shallow KG
embeddings \citep{song_improving_2026,gema_knowledge_2024}. Diffusion- and
flow-matching-based generative models are seeing rapid uptake for
protein backbones and enzyme active sites
\citep{li_flow_2026,ahern_atom-level_2026}, but target single
molecules rather than whole-strain phenotypes. \canopy{} predicts titer from a learned cross-organism representation
that fuses sequence, chemistry, and KG context. The same frozen
embeddings can condition a downstream Bayesian-optimisation loop
\citep{CHENG20232381} or generative model.

\paragraph{Benchmarks for ML in metabolic engineering.}
A useful benchmark for ML-driven metabolic engineering needs three
things: experimental titer measurements, cross-organism coverage, and
multi-omic context for each strain. No existing resource provides all
three. SimDBTL \citep{van_lent_simulated_2023} offers consistent DBTL
splits but its data are simulated rather than experimental.
Therapeutics Data Commons \citep{huang_therapeutics_2021} standardises
prediction splits for therapeutic rather than fermentation tasks.
Text-mining catalogues of over 15{,}000 strain-design publications
\citep{marquez-zavala_database_2025} surface raw records without a
held-out evaluation. \canopy{}'s 4{,}791 fermentation experiments
with a deterministic MD5-hashed 5x CV split address all three
criteria.

\paragraph{Biological knowledge graphs and heterogeneous GNNs.}
Large biomedical KGs (PrimeKG, \citealp{chandak_building_2023}; Hetionet,
\citealp{himmelstein_systematic_2017}; SPOKE, \citealp{nelson_integrating_2019}) and graph
foundation models built on them \citep{huang_foundation_2024,hu_enhancing_2026}
target human disease, not metabolic engineering. We use BioCypher
\citep{lobentanzer_democratising_2023} to integrate ten metabolic-engineering
adapters and build on the Heterogeneous Graph Transformer
\citep{DBLP:journals/corr/abs-2003-01332} with SignNet \citep{lim_sign_2022}, random-walk PE
\citep{dwivedi_graph_2022}, Jumping Knowledge \citep{xu_representation_2018}, and virtual
nodes \citep{gilmer_neural_2017}; pretraining objectives draw on
\citet{hu_strategies_2020} and the graph-FM surveys of
\citet{wang_graph_2025,mao_position_2024}. The closest cross-organism transfer is
IKT4Meta \citep{xin_gene-metabolite_2024}; \canopy{} extends this from
two-organism alignment to a 13-node-type, multi-modal KG with
fermentation-scale supervision.

%% ===================================================================
%%  3  METHOD
%% ===================================================================
\section{Method}

\subsection{Knowledge Graph Construction}
\label{sec:kg}

\canopy{}'s knowledge graph is constructed using BioCypher
\citep{lobentanzer_democratising_2023}. We implement ten adapter modules that ingest
data from complementary sources and yield nodes and edges in a unified
schema.

\paragraph{Node types.}
The graph contains 13 node types spanning five biological scales:
(i)~\emph{molecular} metabolites (InChIKey, SMILES, formula) and
reactions; (ii)~\emph{protein} UniRef90 clusters together with their
InterPro domains; (iii)~\emph{genomic} genes, chassis organisms, and
engineered strains; (iv)~\emph{functional} annotations from the
Gene Ontology, metabolic pathways, and NCBI taxonomy; and
(v)~\emph{experimental} fermentation runs alongside transcriptomic
experiments with per-gene expression measurements.

\paragraph{Edge types.}
Thirty-four edge types capture relationships across these scales:
catalytic associations linking proteins to the reactions they
catalyse, pathway membership, genetic modifications (knockouts,
knockins, and overexpression edits applied to a strain), functional
annotations, ontological structure, experiment tracking, and
gene-expression links between transcriptomic measurements and the
genes they quantify.

\paragraph{Property schema.}
Each node property carries a mandatory prefix that declares its role:
\texttt{sys\_} for system keys used in deduplication (not exported as
features), \texttt{feat\_} for numeric values, \texttt{text\_} for
free text routed to the text encoder, \texttt{seq\_} for amino acid
sequences routed to the protein encoder, and \texttt{smi\_} for SMILES
routed to the chemical encoder. This dispatch ensures graph
construction, embedding, and training share a single source of truth.

\paragraph{Data sources.}
Molecular data is drawn from MetaNetX \citep{10.1093/nar/gkaa992} and KEGG
\citep{kanehisa_kegg_2000}. Protein data comes from UniRef90
\citep{suzek_uniref_2015} with sequences from UniProt
\citep{the_uniprot_consortium_uniprot_2023} and domain annotations from InterPro
\citep{paysan-lafosse_interpro_2023}. Genomic data includes NCBI gene records
\citep{brown_gene_2015} and taxonomy \citep{schoch_ncbi_2020}.
Functional annotations come from the Gene Ontology
\citep{ashburner_gene_2000}. Experimental data is sourced from a
quality-controlled literature corpus and a proprietary experimental
database; transcriptomic data provides per-gene expression linked to
experimental conditions.

\begin{table}[t]
  \caption{Knowledge graph statistics by node type. The graph
  contains 11.2\,M relationships and 17.7\,M node properties in
  total.}
  \label{tab:kg-stats}
  \centering\small
  \begin{tabular}{llr}
    \toprule
    Domain & Node type & Count \\
    \midrule
    \multirow{3}{*}{Genomics}
      & Taxon            & 143 \\
      & Chassis          & 26 \\
      & Genomic Gene     & 168{,}913 \\
    \midrule
    \multirow{2}{*}{Proteomics}
      & UniRef Cluster   & 150{,}914 \\
      & InterPro Domain  & 51{,}489 \\
    \midrule
    \multirow{2}{*}{Metabolomics}
      & Metabolite       & 1{,}495{,}667 \\
      & Reaction         & 83{,}796 \\
    \midrule
    \multirow{4}{*}{Experiment}
      & Strain             & 6{,}910 \\
      & Pathway            & 1{,}872 \\
      & Experiment         & 4{,}791 \\
      & Transcriptomic     & 4{,}860{,}266 \\
    \midrule
    Ontology
      & GO Term (BP/MF/CC) & 38{,}739 \\
    \midrule
    & \textbf{Total}       & \textbf{6{,}863{,}526} \\
    \bottomrule
  \end{tabular}
\end{table}

\subsection{Multi-Modal Feature Encoding}

\canopy{} employs three pretrained foundation models as frozen
extractors. Protein sequences (\texttt{seq\_}) are encoded by ESM-2
\citep{lin_evolutionary-scale_2023} (\texttt{esm2\_t33\_650M\_UR50D}) and mean-pooled
across residues. Chemical structures (\texttt{smi\_}) are encoded by
MoLFormer-XL \citep{ross_large-scale_2021}. Biomedical text
(\texttt{text\_}) is encoded by S-PubMedBERT, applied to GO
definitions, gene names and summaries, pathway names, and experiment
descriptions. Numeric features (\texttt{feat\_}) are z-score
normalised using per-type statistics; categorical integers are
one-hot encoded. All features for a node are concatenated into
$\mathbf{x}_{v}$; per-type input projection handles dimensional
heterogeneity.

\subsection{Model Architecture}

\begin{figure}[t]
  \centering
  \resizebox{\columnwidth}{!}{%
  \begin{tikzpicture}[
    font=\small,
    >={Latex[length=2mm]},
    inp/.style={inner sep=1pt},
    sumc/.style={circle, draw, thick, inner sep=0pt, minimum size=5mm, fill=white},
    box/.style={rounded corners, draw, thick, align=center, inner sep=3pt},
    proj/.style={box, fill=gray!8, minimum width=14mm, minimum height=7mm},
    ntA/.style={circle, draw, thick, minimum size=6mm, inner sep=0pt, fill=blue!15},
    flow/.style={->, thick},
  ]
    \node[inp] (xv)   at (0, 1.4) {$x_v$};
    \node[inp] (eig)  at (0, 0.0) {Eig.\,PE};
    \node[inp] (rw)   at (0,-1.4) {RWPE};
    \node[proj] (nemb)  at (2.6, 1.4) {node\_emb$_{\tau(t)}$};
    \node[proj] (snet)  at (2.6, 0.0) {sign\_net$_{\tau(t)}$};
    \node[proj] (rwemb) at (2.6,-1.4) {rw\_emb$_{\tau(t)}$};
    \draw[flow] (xv.east)  -- (nemb.west);
    \draw[flow] (eig.east) -- node[above, font=\scriptsize] {10-d} (snet.west);
    \draw[flow] (rw.east)  -- node[above, font=\scriptsize] {16-d} (rwemb.west);
    \node[sumc] (sum) at (5.0, 0.0) {$+$};
    \draw[flow] (nemb.east)  to[out=0,in=180] (sum.west);
    \draw[flow] (snet.east)  -- (sum.west);
    \draw[flow] (rwemb.east) to[out=0,in=180] (sum.west);
    \node[ntA] (h0) at (6.4, 0.0) {$h_v^{(0)}$};
    \draw[flow] (sum.east) -- (h0.west);
  \end{tikzpicture}}
  \caption{
    Input feature construction. Each node type $\tau(v)$ has its own
    \textsf{node\_emb}, \textsf{sign\_net} (consuming a 10-d Laplacian
    eigenvector positional encoding), and \textsf{rw\_emb} (consuming a
    16-d random-walk positional encoding); the three streams are summed
    to form the per-node residual stream $h_v^{(0)}$.
  }
  \label{fig:input}
\end{figure}
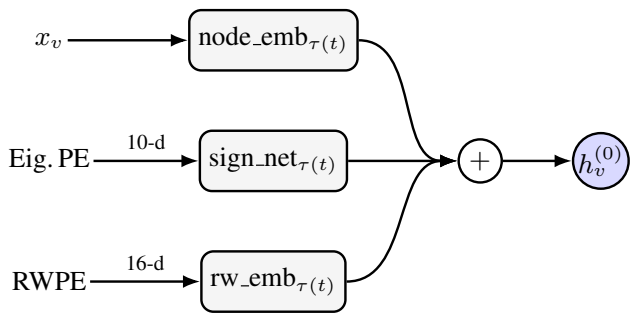

\begin{figure*}[t]
  \centering
  \resizebox{\textwidth}{!}{%
  \begin{tikzpicture}[
    font=\small,
    >={Latex[length=2mm]},
    sumc/.style={circle, draw, thick, inner sep=0pt, minimum size=5mm, fill=white},
    box/.style={rounded corners, draw, thick, align=center, inner sep=3pt},
    proj/.style={box, fill=gray!8, minimum width=14mm, minimum height=7mm},
    attn/.style={box, fill=red!10, minimum width=20mm, minimum height=8mm},
    msg/.style={box, fill=purple!10, minimum width=20mm, minimum height=8mm},
    agg/.style={box, fill=yellow!20, minimum width=13mm, minimum height=7mm},
    outbox/.style={box, fill=cyan!15, minimum width=12mm, minimum height=7mm},
    ntA/.style={circle, draw, thick, minimum size=6mm, inner sep=0pt, fill=blue!15},
    flow/.style={->, thick},
  ]
    \node[ntA] (x) at (0, 0.0) {$t$};
    \node[proj] (Kproj) at (2.4, 1.1) {$K\textsf{-Lin}_{\tau(s)}$};
    \node[proj] (Qproj) at (2.4, 0.0) {$Q\textsf{-Lin}_{\tau(t)}$};
    \node[proj] (Vproj) at (2.4,-1.1) {$V\textsf{-Lin}_{\tau(s)}$};
    \draw[flow] (x.east) to[out=0,in=180] (Kproj.west);
    \draw[flow] (x.east) to[out=0,in=180] (Qproj.west);
    \draw[flow] (x.east) to[out=0,in=180] (Vproj.west);
    \node[attn] (Watt) at (5.6, 0.7) {$K_s\, W^{\textsc{att}}_{\phi}\, Q_t^\top /\sqrt{d}$ \\[-1pt]
      {\scriptsize $\cdot\, \mu_{\langle\tau(s),\phi,\tau(t)\rangle}$}};
    \node[msg]  (Wmsg) at (5.6,-1.1) {$V_s\, W^{\textsc{msg}}_{\phi}$};
    \draw[flow] (Kproj.east) to[out=0,in=180] ([yshift=2mm]Watt.west);
    \draw[flow] (Qproj.east) to[out=0,in=180] ([yshift=-2mm]Watt.west);
    \draw[flow] (Vproj.east) to[out=0,in=180] (Wmsg.west);
    \node[box, fill=red!5, minimum width=18mm, minimum height=7mm] (sm) at (8.6, 0.7)
         {$\mathrm{softmax}_{s\in\mathcal{N}(t)}$};
    \draw[flow] (Watt.east) to[out=0,in=180] (sm.west);
    \node[agg] (agg) at (8.6,-1.1) {$\displaystyle\sum_{s} \alpha_{s\to t}\, m_{s\to t}$};
    \draw[flow] (sm.south) to[out=-90,in=90] (agg.north);
    \draw[flow] (Wmsg.east) to[out=0,in=180] (agg.west);
    \node[outbox] (out) at (11.0,-1.1) {$A\textsf{-Lin}_{\tau(t)}$};
    \draw[flow] (agg.east) to[out=0,in=180] (out.west);
    \node[sumc] (sum1) at (12.8,-1.1) {$+$};
    \draw[flow] (out.east) to[out=0,in=180] (sum1.west);
    \node[box, fill=cyan!8, minimum width=22mm, minimum height=12mm,
          font=\scriptsize] (ffn) at (15.0,-1.1)
         {LayerNorm \\ Linear \\ GELU \\ Dropout \\ Linear};
    \draw[flow] (sum1.east) to[out=0,in=180] (ffn.west);
    \node[sumc] (sum2) at (17.2,-1.1) {$+$};
    \draw[flow] (ffn.east) to[out=0,in=180] (sum2.west);
    \node[ntA] (tnew) at (18.6,-1.1) {$t'$};
    \draw[flow] (sum2.east) to[out=0,in=180] (tnew.west);
    \draw[flow, blue!60!black, rounded corners]
       (x.north) -- (x.north |- 0,2.0) --
       node[above, font=\scriptsize] {Attention residual}
       (sum1.north |- 0,2.0) -- (sum1.north);
    \node[font=\scriptsize, blue!60!black, fill=white, inner sep=1pt]
         at ($(sum1.north)+(0,0.55)$)
         {$\times\,\alpha^{(\text{att})}_{[\tau]}$};
    \draw[flow, blue!60!black, rounded corners]
       (sum1.south) -- (sum1.south |- 0,-2.6) --
       node[below, font=\scriptsize] {FFN residual}
       (sum2.south |- 0,-2.6) -- (sum2.south);
    \node[font=\scriptsize, blue!60!black, fill=white, inner sep=1pt]
         at ($(sum2.south)+(0,-0.55)$)
         {$\times\,\alpha^{(\text{ffn})}_{[\tau]}$};
    \node[below=1pt of tnew, font=\scriptsize] {$h_t^{(\ell+1)}$};
  \end{tikzpicture}}
  \caption{
    One HGT layer. The dst-side query $Q_{\tau(t)}$ is type-specific;
    keys $K_{\tau(s)}$ and values $V_{\tau(s)}$ are projected per
    source type, while attention logits and messages are scaled by
    per-relation matrices $W^{\textsc{att}}_{\phi}, W^{\textsc{msg}}_{\phi}$
    and a learnable relation prior
    $\mu_{\langle\tau(s),\phi,\tau(t)\rangle}$. Attention is softmaxed
    over the destination's neighbourhood and aggregated; a per-type
    output projection $A\textsf{-Lin}_{\tau(t)}$ feeds an FFN sub-block.
    Both attention and FFN use learnable scaled residual connections (blue).
  }
  \label{fig:hgt}
\end{figure*}
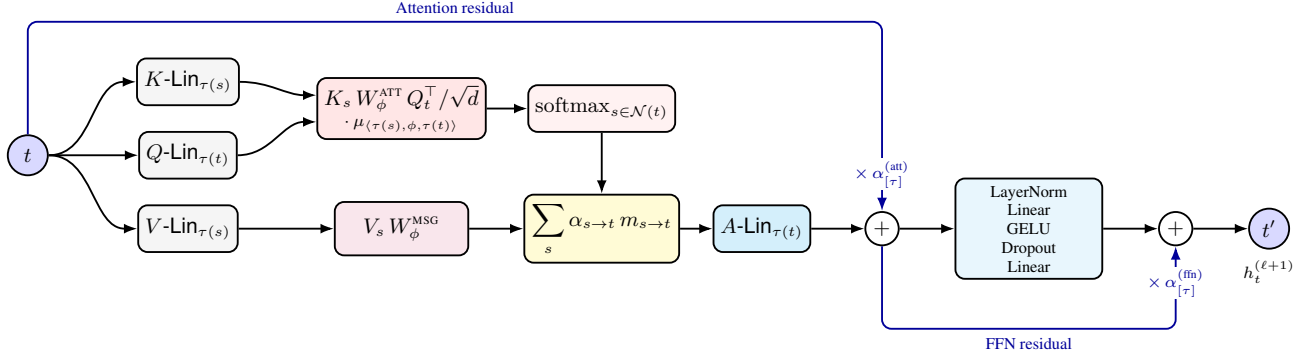

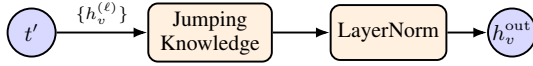
\begin{figure}[t]
  \centering
  \resizebox{0.85\columnwidth}{!}{%
  \begin{tikzpicture}[
    font=\small,
    >={Latex[length=2mm]},
    box/.style={rounded corners, draw, thick, align=center, inner sep=3pt},
    proj/.style={box, fill=orange!12, minimum width=18mm, minimum height=8mm},
    ntA/.style={circle, draw, thick, minimum size=7mm, inner sep=0pt, fill=blue!15},
    flow/.style={->, thick},
  ]
    \node[ntA] (tnew) at (0,0) {$t'$};
    \node[proj] (jk) at (2.6,0) {Jumping\\Knowledge};
    \node[proj, minimum width=14mm] (lnout) at (5.2,0) {LayerNorm};
    \node[ntA] (hout) at (7.0,0) {$h_v^{\mathrm{out}}$};
    \draw[flow] (tnew.east) -- node[above, font=\scriptsize]
                {$\{h_v^{(\ell)}\}$} (jk.west);
    \draw[flow] (jk.east) -- (lnout.west);
    \draw[flow] (lnout.east) -- (hout.west);
  \end{tikzpicture}}
  \caption{
    Readout. After $L$ stacked HGT layers, JumpingKnowledge
    max-pools the per-layer representations $\{h_v^{(\ell)}\}$
    (combined with a learnable input-residual scale
    $\alpha^{\text{(in)}}_{[\tau]}$) and a final per-type LayerNorm
    produces the encoder output $h_v^{\mathrm{out}}$.
  }
  \label{fig:readout}
\end{figure}

\begin{figure*}[t]
  \centering
  \resizebox{\textwidth}{!}{%
  \begin{tikzpicture}[
    font=\small,
    >={Latex[length=2mm]},
    box/.style={rounded corners, draw, thick, align=center, inner sep=4pt},
    head/.style={box, fill=orange!12, minimum width=58mm, minimum height=11mm,
                 align=left},
    enc/.style={box, fill=blue!15, minimum width=30mm, minimum height=18mm,
                align=center},
    comb/.style={box, fill=cyan!15, minimum width=58mm, minimum height=22mm,
                 align=center},
    flow/.style={->, thick},
  ]
    \node[enc] (z) at (0, -3.0) {$\{z_v\}_{v\in V}$ \\ encoder output};

    \node[head, anchor=west] (h1) at (3.6, -0.5)
      {\textbf{Link prediction} \\
       $\sigma(\mathrm{MLP}(z_u, z_v)) \to$ BCE};
    \node[head, anchor=west] (h2) at (3.6, -2.2)
      {\textbf{Masked node model} \\
       $\mathsf{mnm\_head}_{[\tau]}(z_v) \to$ MSE on $x_v$};
    \node[head, anchor=west] (h3) at (3.6, -3.9)
      {\textbf{Graph distance} \\
       $\mathrm{MLP}(z_i \odot z_j) \to$ MSE (shortest-path)};
    \node[head, anchor=west] (h4) at (3.6, -5.6)
      {\textbf{Contrastive Exp$\leftrightarrow$Exp} \\
       $\cos(z_i, z_j) \to$ MSE on $1 - d/d_{\max}$};

    \node[comb, anchor=west] (L) at (12.6, -3.0)
      {$\mathcal{L} = \displaystyle\sum_i \tfrac{\mathcal{L}_i}{2\sigma_i^2} + \log\sigma_i$ \\[2pt]
       {\scriptsize Kendall \& Gal, 2018 (uncertainty weighting)}};

    \draw[flow] (z.east) to[out=0,in=180] (h1.west);
    \draw[flow] (z.east) to[out=0,in=180] (h2.west);
    \draw[flow] (z.east) to[out=0,in=180] (h3.west);
    \draw[flow] (z.east) to[out=0,in=180] (h4.west);

    \draw[flow] (h1.east) to[out=0,in=180] (L.west);
    \draw[flow] (h2.east) to[out=0,in=180] (L.west);
    \draw[flow] (h3.east) to[out=0,in=180] (L.west);
    \draw[flow] (h4.east) to[out=0,in=180] (L.west);
  \end{tikzpicture}}
  \caption{
    Pretraining objective. Four parallel heads consume the encoder
    output $\{z_v\}_{v\in V}$: link prediction (BCE), per-type masked
    node modelling (MSE), graph-distance regression (MSE), and an
    Experiment$\leftrightarrow$Experiment contrastive loss (MSE on
    cosine similarity). Their weighted sum, using \citet{kendall_multi-task_2018}
    uncertainty weighting, forms the training objective.
  }
  \label{fig:losses}
\end{figure*}
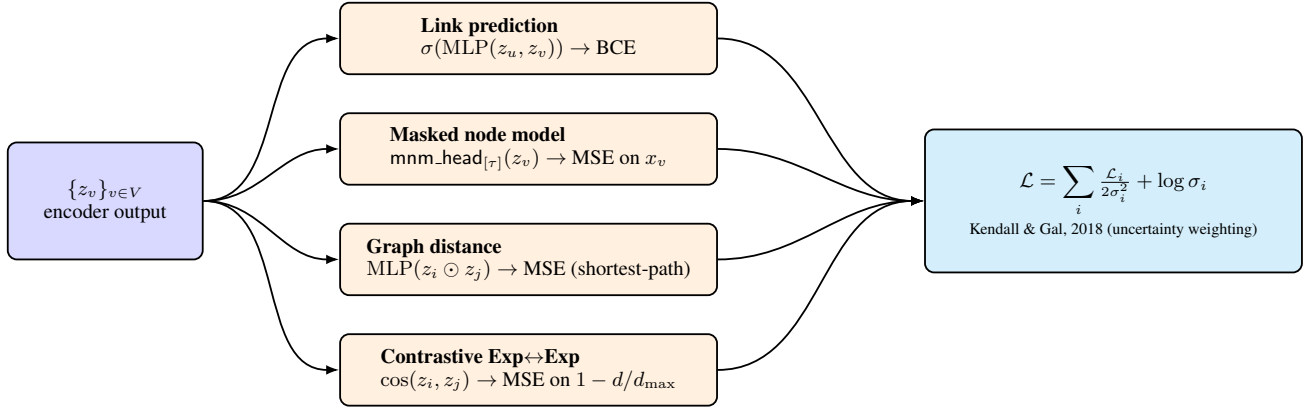

\canopy{}'s encoder is a Heterogeneous Graph Transformer that
processes typed nodes and edges natively, extending HGTConv
\citep{DBLP:journals/corr/abs-2003-01332} with several modern components. Figures~\ref{fig:input}--\ref{fig:readout}
give a schematic overview of the full architecture, including the
per-type and per-relation parameterisation that distinguishes HGT
from a homogeneous GNN.

\paragraph{Input projection.}
For each node type~$t$, a learnable linear projection
$\mathbf{W}_{t}$ maps $\mathbf{x}_{v}$ to a hidden representation of
dimension $d = d_{h} \cdot H$, with $d_{h}$ the per-head dimension and
$H$ the number of heads.

\paragraph{Positional encodings.}
We augment node features with two complementary signals. The $k$
smallest non-trivial Laplacian eigenvectors are computed via
randomised SVD and processed through SignNet
\citep{lim_sign_2022}:
\begin{equation}
  \text{SignNet}(\mathbf{e}) = \text{MLP}(\mathbf{e}) + \text{MLP}(-\mathbf{e}).
\end{equation}
Random walk structural encodings of length~$\ell$ are computed per subgraph sample
and projected per-type. Both encodings are
added element-wise to the projected features.

\paragraph{Transformer blocks.}
The model stacks $L$ blocks: per-type pre-norm; HGTConv with
type-dependent multi-head attention; GELU; a residual connection
scaled by a learnable \textsc{ScaleLayer} initialised at 0.1; a
per-type FFN with expansion $r$ and dropout; and a second scaled
residual. The small initial scale stabilises training of deep
networks by letting the residual stream dominate early.
On top of the stack, we apply Jumping Knowledge \citep{xu_representation_2018} with max-pooling across
all $L$ layers per type. A virtual node type is added with
bidirectional edges to all other nodes, reducing effective graph
diameter and providing a shortcut path between otherwise disconnected
components. The final representations are layer-normalised.

\subsection{Multi-Task Self-Supervised Pretraining}

We pretrain \canopy{} with four complementary objectives.

\paragraph{Hold-out integrity.}
The same hash-keyed Experiment-node split used at probe time
(Section~\ref{sec:setup}) gates every pretraining objective that touches an
\texttt{Experiment} node: edges incident to any val/test Experiment
are excluded from the link-prediction label set, masked-node-modelling
ignores held-out Experiment rows when sampling its mask, and the
contrastive Experiment-clustering loss filters held-out Experiments
out of its pair pool before sampling. Held-out Experiment nodes
remain in the message-passing graph so probe-time message passing
matches the pretrain-time graph the model was conditioned on; only
\emph{supervision} signal involving their identity, features, or
neighbour structure is removed.

\paragraph{Link prediction.}
Thirty percent of edges are deterministically reserved as supervision
labels via an MD5 hash of edge endpoints, ensuring consistent splits
across sampling runs (subject to the hold-out filter above).
A dot-product predictor scores edges,
$\hat{y}_{uv} = \sigma(\mathbf{z}_{u}^{\top}\mathbf{z}_{v})$, trained
with binary cross-entropy. For each edge type we draw negatives at a
1:1 ratio with positives by uniformly sampling type-constrained
$(src, dst)$ pairs from the corresponding node pools within the
sampled subgraph and rejecting pairs that are already edges.

\paragraph{Masked node modelling.}
We mask 15\% of node features per batch. Per-type linear decoders
reconstruct masked features under MSE, which acts as a regulariser
against representation collapse.

\paragraph{Distance prediction.}
For each subgraph, 200 random node pairs are sampled and their
shortest-path distances computed via BFS on the undirected graph
before virtual-node edges are added (capped at 5 hops); virtual
nodes would otherwise collapse the diameter to two and trivialise
the target. A distance head (the Hadamard product of source and
target embeddings followed by a two-layer MLP) regresses these
distances under MSE, supervising pairwise embedding distances
against the graph metric.

\paragraph{Contrastive experiment-pair loss.}
Experiment node embeddings are contrasted within each batch with a
target similarity inversely proportional to graph distance,
$s_{ij} = 1 - d_{ij}/d_{\max}$, trained with temperature-scaled
cosine similarity ($\tau {=} 0.07$).

\paragraph{Loss combination.}
The four losses are combined via homoscedastic uncertainty weighting
\citep{kendall_multi-task_2018}. Each task has a learnable log-variance
$\log\sigma_{i}$, and the total loss is
\begin{equation}
  \mathcal{L} = \sum_{i}
    \frac{\mathcal{L}_{i}}{2\sigma_{i}^{2}} + \log\sigma_{i}.
\end{equation}
$\log\sigma_{i}$ is clamped at $-4.0$ to prevent collapse; tasks with
noisier supervision automatically receive lower weight.

\subsection{Scalable Subgraph Sampling}
\label{sec:sampling}

Training on the full graph is infeasible; \canopy{} uses a
Neo4j-backed streaming sampler that constructs mini-batch subgraphs
in parallel worker processes and writes each to disk as a PyTorch
Geometric \citep{fey_fast_2019} \texttt{HeteroData} object. The sampler
supports three strategies (simple BFS, batched random walks, and
multi-anchor); we use multi-anchor by default. Seeds are split
50/50 between yield-bearing Experiment nodes (so every probe target
appears in some subgraph) and inverse-degree-weighted random nodes
(so peripheral nodes are not crowded out by hubs).

For each Experiment seed batch the multi-anchor strategy resolves
four BFS anchors: the seed Experiment; its \textsc{Strain} (via
\textsc{uses\_strain}); parent \textsc{Taxon} plus capped sibling
strains (via \textsc{belongs\_to}, capped at 10 per species); and
adjacent \textsc{MetabolicPathway}s (via \textsc{has\_pathway} from
the Strain). It also resolves four directed metapath anchors that
walk fixed causal chains the undirected BFS under-covers:
Experiment$\to$\textsc{Metabolite} (target compound),
Pathway$\to$Reaction$\to$\textsc{UnirefCluster} (pathway enzymes),
Strain$\to$\textsc{GenomicGene}$\to$\textsc{UnirefCluster} (genetic
edits to encoded proteins), and the reverse-direction chain
Metabolite$\to$Reaction$\to$\textsc{UnirefCluster}$\to$\textsc{GenomicGene}
(target enzymes). Each anchor is allocated a fraction of a global
node budget $N_{\max}$ (default 1{,}000); unused budget from inactive
anchors is proportionally redistributed. Ontological edges
(\textsc{subclass\_of}, \textsc{part\_of}, \textsc{regulates}) and
transcriptomic measurement edges are excluded from BFS expansion to
prevent hub-dominated subgraphs but are included in the final
subgraph if both endpoints are present.
The multi-anchor configuration rebalances the GenomicGene fraction
from 78\% to 15\% and increases Experiment-node coverage roughly
$13\times$ relative to a naive 5-hop BFS. The 30\% link-supervision
split is computed deterministically via MD5 hashing of
\texttt{(src, rel, dst)}, so an edge belongs to the same set
regardless of which subgraph contains it; edges incident to held-out
Experiment nodes are kept in the message-passing graph but excluded
from supervision. Continuous scalar features are z-score normalised
per node type, and random-walk positional encodings are added to
each subgraph at materialisation.

\subsection{Downstream Tasks}

The prediction target is the measured product titer of each
fermentation Experiment: the experimentally reported concentration of
the target compound from a physical microbial culture. Titers are
drawn from two sources: published metabolic-engineering studies
(literature-mined) and unpublished in-house design-build-test-learn
runs recorded in our LIMS.
After pretraining, we train lightweight probes on frozen
experiment-node embeddings: a linear probe and a two-layer MLP probe
for titer regression. We report $R^{2}$, RMSE, and
Spearman~$\rho$ for regression, and AUROC and F1 for binary
classification (above/below median titer). This protocol isolates
representation quality from probe capacity.

%% ===================================================================
%%  4  EXPERIMENTS
%% ===================================================================
\section{Experiments}

\subsection{Setup}
\label{sec:setup}

\paragraph{Knowledge graph statistics.}
\canopy{}'s graph comprises 6.9\,M nodes across 13 types and 11.2\,M
edges across 34 typed relations (Table~\ref{tab:kg-stats},
Section~\ref{sec:kg}).
At training time we additionally materialise reverse edges,
self-loops, and virtual-node edges, bringing the per-batch
typed-tuple count above 100.

\paragraph{Sampling.}
We sample 10{,}000 subgraphs with the multi-anchor expansion
($k{=}3$, $N_{\max}{=}1000$) of Section~\ref{sec:sampling}. Pretraining
uses the deterministic MD5 edge-supervision split described in
Section~\ref{sec:sampling}; the downstream titer probe uses a 5x CV
Experiment-node split (Experiment IDs hashed and bucketed) so test
experiments are unseen during both pretrain and probe fitting.

\paragraph{Model configurations.}
We evaluate three scales:

\begin{table}[h]
  \caption{Model configurations used in this paper.}
  \label{tab:configs}
  \centering\small
  \begin{tabular}{lccccc}
    \toprule
    Config & Hidden & Heads & Layers & FFN & Params \\
    \midrule
    Demo & 64  & 4 & 6  & 4 & $\sim$80M \\
    500M & 128 & 4 & 6  & 4 & $\sim$0.5B \\
    3B   & 256 & 8 & 12 & 4 & $\sim$3B   \\
    \bottomrule
  \end{tabular}
\end{table}

\paragraph{Training.}
AdamW ($\beta_{1}{=}0.9, \beta_{2}{=}0.999$) with cosine annealing and
linear warmup; bfloat16 mixed precision; FSDP for distributed
training across Intel Data Center GPU
Max accelerators (Dawn). Gradient clipping at $\ell_{2}$ norm 1.0.
Hyperparameters tuned via a two-tier Optuna \citep{akiba_optuna_2019} sweep: Tier~1
(1{,}000 trials at 500M, $\sim$500 XPU-hours, search over
$h \in \{64,128,192,256\}$, $L \in \{4,6,8\}$); Tier~2 (200 trials
at 3B, $\sim$800 XPU-hours, search narrowed from Tier~1).
A single model trains to convergence in 7~XPU-hours at 500M (7~h on one
Max GPU) and 160~XPU-hours at 3B (40~h on four); the $\sim$500 and
$\sim$800 XPU-hour budgets above amortise over the 1{,}000 and 200
median-pruned trials of each tier. Reported parameter counts are for the
trainable HGT only: the ESM-2, MoLFormer, and PubMedBERT encoders are
frozen and their node features precomputed, so pretraining does not
backpropagate through them.

\paragraph{Baselines.}
\emph{Ridge}, \emph{MLP} (two hidden layers, 128$\to$64), and
\emph{XGBoost} (500 trees, depth 6) trained directly on the 429-dim
raw experiment-condition feature vector (the same input that
feeds the Experiment node in the graph), using the identical
MD5-hashed train/test split as the \canopy{} probe.  We further
compare against graph baselines sharing the same probe and split: a
homogeneous \emph{GraphSAGE} backbone (evaluated both in a
\emph{vanilla} form and with \canopy{}'s SignNet, Jumping Knowledge, and
virtual-node augmentations) and a \emph{vanilla HGT} backbone without
those augmentations. Together with \canopy{} itself (the HGT backbone
\emph{with} the augmentations), these separate the contribution of the
heterogeneous backbone from that of the augmentations.

\subsection{Main Results}

\begin{table}[t]
  \caption{Titer prediction on the held-out test split
    ($n{=}410$ unique experiments). All methods share the same
    MD5-hashed split. The graph baselines separate backbone
    (homogeneous SAGEConv vs.\ heterogeneous HGT) from \canopy{}'s
    SignNet/JK/virtual-node augmentations (SN/JK/VN); \canopy{} is the
    HGT backbone with those augmentations.}
  \label{tab:main-results}
  \centering\small
  \begin{tabular}{lcc}
    \toprule
    Method & $R^{2}\uparrow$ & AUROC$\uparrow$ \\
    \midrule
    Ridge (raw conditions)   & 0.086 & 0.698 \\
    XGBoost (raw conditions) & 0.236 & 0.726 \\
    MLP (raw conditions)     & 0.078 & 0.691 \\
    GraphSAGE (vanilla)      & 0.241 & 0.713 \\
    GraphSAGE $+$ SN/JK/VN    & 0.334 & 0.734 \\
    HGT (vanilla)            & 0.308 & 0.711 \\
    \midrule
    \canopy{} (Demo)         & 0.301 & 0.703  \\
    \canopy{} (500M)         & 0.380 & 0.805  \\
    \canopy{} (3B)           & 0.413  & 0.820  \\
    \bottomrule
  \end{tabular}
\end{table}

\subsection{Ablations}

Unless stated otherwise, ablations are run at the 500M scale defined
in Section~\ref{sec:setup} with a 5k-sample budget; the
runs in Table~\ref{tab:main-results} use the full 10k-sample
budget. Absolute $R^{2}$ values in the ablations therefore sit below
the headline numbers and should be read as relative comparisons.

We ablate the four self-supervised tasks
(Table~\ref{tab:loss-ablation}), the three architectural
augmentations (SignNet PE, Jumping Knowledge, and virtual nodes;
Table~\ref{tab:arch-ablation}), and network depth
(Table~\ref{tab:depth-ablation}). For depth, we compare the default
$L{=}6$ architecture against an iso-parameter shallow variant
($L{=}2$, $h{=}224$) at 500M scale. The shallow model reaches a
higher peak probe $R^{2}$, runs $\sim$2$\times$ faster per epoch,
and, unlike the deeper variant, does not exhibit probe degradation
through 20 epochs. We attribute this to a lower oversmoothing burden
once HGTConv attention has enough per-head capacity ($d_{h}{=}56$);
Jumping Knowledge alone is insufficient at $L{=}6$ on this graph. We
retain the deeper default in the headline configuration: the gap is
small ($\Delta R^{2}{=}0.011$) and the deeper topology matches the
3B configuration from which we report headline numbers. Mitigations
beyond JK and scaled residuals (DropEdge, GraphNorm, or deeper
feature fusion) remain unexplored and are a natural next ablation.

\begin{table}[t]
  \caption{Pretraining-objective ablation (probe $R^{2}$, 500M
    scale, 5k samples).}
  \label{tab:loss-ablation}
  \centering\small
  \begin{tabular}{cccccc}
    \toprule
    Link & MNM & Dist & Contrast & $R^{2}\uparrow$ / $\Delta$pp \\
    \midrule
    \checkmark & \checkmark & \checkmark & \checkmark & \textbf{0.359} \\
               & \checkmark & \checkmark & \checkmark & $-1.6$ \\
    \checkmark &            & \checkmark & \checkmark & $-0.6$ \\
    \checkmark & \checkmark &            & \checkmark & $-4.7$ \\
    \checkmark & \checkmark & \checkmark &            & $-7.6$ \\
    \bottomrule
  \end{tabular}
\end{table}

\begin{table}[t]
  \caption{Architectural ablation (500M, 5k samples; probe $R^{2}$).}
  \label{tab:arch-ablation}
  \centering\small
  \begin{tabular}{lc}
    \toprule
    Variant & $R^{2}\uparrow$ / $\Delta$pp \\
    \midrule
    Full \canopy{}         & \textbf{0.359} \\
    $-$ SignNet PE         & $-2.7$ \\
    $-$ JK aggregation     & $-3.5$ \\
    $-$ Virtual nodes      & $-10.3$ \\
    $-$ All three          & $-11.1$ \\
    \bottomrule
  \end{tabular}
\end{table}

\begin{table}[t]
  \caption{Depth ablation (iso-parameter, 500M scale, 5k samples).}
  \label{tab:depth-ablation}
  \centering\small
  \begin{tabular}{lcc}
    \toprule
    Variant & $R^{2}\uparrow$ / $\Delta$pp & AUROC / $\Delta$pp \\
    \midrule
    $L{=}2,\,h{=}224$ (shallow) & \textbf{0.3703} & \textbf{0.7881} \\
    $L{=}6,\,h{=}128$ (deep) & $-1.0$ & $-1.1$ \\
    \bottomrule
  \end{tabular}
\end{table}

\paragraph{Task weighting.}
The four pretraining losses operate on very different scales: BCE for
link prediction, MSE for masked-node and distance, and contrastive
cosine for experiment pairs. Under flat (equal) weighting the
highest-scale loss dominates the gradient and the probe collapses:
removing the learned weighting drops $R^{2}$ by $9.18$
($\Delta R^{2}{=}{-}9.18$ relative to the learned scheme), well below
the constant-mean predictor (Table~\ref{tab:weighting}). The
homoscedastic uncertainty scheme of \citet{kendall_multi-task_2018}
attains $R^{2}{=}0.359$. The learned $\log\sigma_{i}$ values
converge to per-task scales that balance gradient contributions
automatically and adapt as the relative difficulty of each task
shifts during training, removing the manual sweep over fixed
per-task weights that would otherwise be needed.

\begin{table}[t]
  \caption{Task-weighting comparison (500M, 5k samples).}
  \label{tab:weighting}
  \centering\small
  \begin{tabular}{lcc}
    \toprule
    Weighting & $R^{2}\uparrow$ & $\Delta R^{2}$ \\
    \midrule
    Learned (Kendall) & 0.359 & --- \\
    Flat (equal)      & ---   & $-9.18$ \\
    \bottomrule
  \end{tabular}
\end{table}

%% ===================================================================
%%  5  DISCUSSION
%% ===================================================================
\section{Discussion}

Each of the four pretraining objectives contributes
(Table~\ref{tab:loss-ablation}), with no single task dominating.
Learned uncertainty weighting
(Table~\ref{tab:weighting}) eliminates manual loss balancing and
adapts as training progresses.

The 3B model improves $R^{2}$ from 0.380 to 0.413 over the 500M
model, a modest gain relative to the 6$\times$ parameter increase.
With 4{,}791 total literature-mined fermentation records, the regime is likely
data-bound rather than capacity-bound at this scale; substantiating
scaling claims will require more experimental data, particularly
across the long tail of compounds and organisms.
In deployment, titer prediction is used to rank candidate designs for
the wet lab rather than to replace measurement; in this triage setting
the meaningful gain is in regression, from the best tabular baseline
($R^{2}{=}0.24$) to \canopy{}'s cross-organism representation
($R^{2}{=}0.41$). We report AUROC for completeness but read it with
caution: fermentation volume (\texttt{feat\_volume}) correlates with
titer (Pearson $0.15$) and on its own separates above/below-median
titer at AUROC $0.65$, so the binary task is partly trivialised and all
methods sit in a narrow AUROC band regardless of regression skill.
Because each fermentation costs weeks of bench time, even a modest
improvement in how designs are ranked can reduce the number of physical
builds per successful design, though we do not yet quantify this
end-to-end.

\paragraph{Limitations.}
First, experimental data is sparse relative to the KG, with uneven
distribution across compounds and organisms. Second, the model
is predictive rather than mechanistic; attention analysis offers
some interpretability but does not replace mechanistic modelling.
Third, our baselines do not include a graph-free pooled-encoder
control, e.g., concatenating frozen ESM-2 and MoLFormer embeddings
of a strain's genes and target compound and applying an MLP; such
a probe would isolate the contribution of graph structure from the
underlying foundation-model encoders and is a planned ablation.
Fourth, release scope is constrained for this submission: the
4{,}791-experiment literature-mined benchmark and its split files
will be released in a forthcoming publication, while the in-house
LIMS records and trained model weights are not released with this
workshop paper, with an archival release of the full pipeline
planned to accompany a later journal submission. Fifth, we do not yet
isolate the contribution of individual data sources. A node-type
feature-masking ablation, which zeros proteomic (UniRef/InterPro),
genomic, or transcriptomic features at probe time, would reveal which
modalities are influential versus redundant. A per-fold check would
test whether the experiment-level split induces feature-distribution
shift. Both are planned and inform the data-expansion priorities below.

\paragraph{Future work.}
Several extensions are already underway. On the representation side,
we are scaling node features with larger and more sophisticated
embedding models, ingesting orders of magnitude more protein
sequences, and adding predicted protein structures as a complementary
modality. On the data side, we are expanding the KG with DNA parts
(promoters, RBSs, terminators, CDS variants) and metagenomic
sequences from environmental and engineered communities, broadening
coverage well beyond the current cultured-organism backbone. On the
application side, the same frozen embeddings are being extended from
titer prediction to chassis selection and de novo pathway design,
both of which reuse \canopy{}'s heterogeneous representation but
condition on different query node types. Substantiating the
``foundation model'' framing also requires breadth in downstream
evaluation; beyond titer prediction, we plan probes on chassis
selection, gene essentiality, reaction prediction, and
cross-organism transfer (e.g., pretraining on \emph{E.~coli}
experiments and evaluating on yeast) to test whether the same
frozen embeddings transfer across distinct metabolic-engineering
tasks.
Finally, we are pairing the learned representation with generative
strain-design pipelines: a flow-matching model conditioned on
\canopy{} embeddings proposes candidate multi-gene interventions,
which are then scored by the frozen titer probe inside a
Bayesian-optimisation loop, turning the predictive oracle into a
closed-loop generative design system.

\paragraph{Broader impact.}
Accelerating strain design supports a shift from petrochemical to
fermentative production of chemicals, fuels, and therapeutics.

%% ===================================================================
%%  6  CONCLUSION
%% ===================================================================
\section{Conclusion}

Frozen \canopy{} embeddings outperform tabular and homogeneous graph
baselines on titer prediction, providing a cross-organism
representation that can be reused as the conditioning substrate for
downstream generative strain-design pipelines.

%% ===================================================================
%%  IMPACT STATEMENT  (ICML required)
%% ===================================================================
\section*{Impact Statement}
This paper presents work whose goal is to advance the field of
machine learning for biological design. There are many potential
societal consequences of our work, including positive impact through
greener bioproduction.

%% ===================================================================
%%  ACKNOWLEDGEMENTS
%% ===================================================================
\section*{Acknowledgements}
The authors acknowledge the use of resources provided by the Dawn
National AI Research Resource (AIRR). Dawn is operated by the
University of Cambridge and is funded by the UK Government's Department
for Science, Innovation and Technology (DSIT) via UK Research and
Innovation, the Science and Technology Facilities Council
[ST/Z000890/1], Dell Technologies and Intel.

%% ===================================================================
%%  REFERENCES
%% ===================================================================
\FloatBarrier
\bibliographystyle{icml2026}

\bibliography{canopy_genbio}

%% ===================================================================
%%  APPENDIX
%% ===================================================================
\newpage
\appendix
\onecolumn

\section{Knowledge graph construction details}
\label{app:kg-details}

\canopy{}'s graph is assembled with BioCypher \citep{lobentanzer_democratising_2023},
which separates \emph{what} the graph contains from \emph{how} it is
collected. A single declarative schema (YAML, keyed by Biolink-aligned
node and edge types) fixes the allowed types, their identifier
namespaces, and their property contracts. Ten adapter modules each emit
typed node and edge streams against that schema; BioCypher rejects
labels that are not declared, collapses sub-classes onto the declared
parent type, and writes \texttt{neo4j-admin}-importable CSVs. Adapters
can therefore be added, swapped, or version-pinned without touching
graph code, and the same schema drives both build-time validation and
downstream sampling. The ten adapters span public reference resources
(sequence, structure, pathway, ontology), a literature-mined
fermentation corpus, and one in-house LIMS adapter contributing
unpublished DBTL records (Table~\ref{tab:adapters}).

\begin{table}[h]
\caption{BioCypher adapters used to construct \canopy{}'s KG.}
\label{tab:adapters}
\centering\small
\begin{tabular}{lp{4.2cm}p{8cm}}
\toprule
Adapter & Source & Nodes / edges produced \\
\midrule
Genomic              & UniProt REST + curated chassis list & \texttt{Chassis}, \texttt{GenomicGene}; \texttt{ENCODES} (Gene$\to$UnirefCluster), \texttt{HAS\_GENE} (Chassis$\to$Gene). \\
UniRef               & UniRef90 \texttt{.tsv}, MetaNetX cross-refs & \texttt{UnirefCluster}; \texttt{CATALYZED\_BY} (Reaction$\to$UnirefCluster), \texttt{HAS\_GO\_TERM}. \\
InterPro             & \texttt{protein2ipr.dat.gz} (bulk) or UniProt batch API & \texttt{InterProDomain}; \texttt{HAS\_DOMAIN} (UnirefCluster$\to$Domain). \\
Gene Ontology        & GO OBO release (\texttt{obonet}) & \texttt{GOTerm} (BP/MF/CC); ontology hierarchy edges (\textsc{subclass\_of}, \textsc{part\_of}, \textsc{regulates}). \\
MetaNetX             & MNXref \citep{10.1093/nar/gkaa992} & \texttt{Metabolite}, \texttt{Reaction}; \texttt{HAS\_PARTICIPANT}, \texttt{HAS\_PRODUCT}. \\
Taxonomy             & NCBI Taxonomy & \texttt{Taxon}; \texttt{BELONGS\_TO} (Strain$\to$Taxon). \\
Transcriptomic       & Public RNA-seq compendia & \texttt{Transcriptomic} measurement nodes; \texttt{MEASURED\_BY\_TRANSCRIPTOMIC}, \texttt{DERIVED\_FROM\_TRANSCRIPTOMIC}, \texttt{USES\_STRAIN\_TRANSCRIPTOMIC} edges. \\
Experimental scrape  & Literature-mined fermentation records & \texttt{Experiment}, \texttt{Strain}, \texttt{MetabolicPathway}; \texttt{USES\_STRAIN}, \texttt{HAS\_PATHWAY}, \texttt{TARGETS\_COMPOUND}, \texttt{MEASURES\_COMPOUND}, \texttt{HAS\_KNOCKOUT}, \texttt{HAS\_KNOCKIN}, \texttt{HAS\_OVEREXPRESSION}. \\
Pathway enrichment   & KEGG REST & enriches \texttt{MetabolicPathway} nodes with KEGG metadata; \texttt{HAS\_STEP} (Pathway$\to$Reaction). \\
LIMS (in-house)      & Internal experiment registry & supplements \texttt{Experiment} / \texttt{Strain} with unpublished DBTL records and the edit-edge types above. \\
\bottomrule
\end{tabular}
\end{table}

\paragraph{Schema harmonisation.}
BioCypher's ontology layer maps every adapter's emitted labels to
Biolink superclasses. Only types declared in the project schema are
retained at build time, and sub-class collapse prevents label
fragmentation (e.g.\ ``BiologicalProcess'' and ``MolecularFunction''
are both rolled up under \texttt{GOTerm}). Edges referencing undeclared
node types are dropped before import and logged.

\paragraph{Identifier normalisation and deduplication.}
Cross-source identifiers follow a fixed precedence: MetaNetX MNX IDs
for metabolites and reactions, UniRef90 cluster IDs for proteins, NCBI
gene IDs for genes, NCBI Taxonomy IDs for taxa, GO IDs for ontology
terms, and InterPro IDs for domains. Organism names in the
literature-mined corpus are normalised through a curated synonym table
that maps frequently-confused taxonomic labels (e.g.\ \emph{Pichia
pastoris}~$\to$~\emph{Komagataella phaffii}, \emph{Clostridium
thermocellum}~$\to$~\emph{Acetivibrio thermocellus}) to the canonical
chassis list. Duplicate metabolite and reaction nodes from MetaNetX
cross-references are merged on MNX ID; \texttt{Uniref$\to$Reaction} and
\texttt{Uniref$\to$GOTerm} mappings are resolved against the MetaNetX
curated reaction model and the GO release pinned at build time. Edges
with malformed or missing endpoints are dropped at adapter time and
logged.

\paragraph{Build pipeline.}
Adapters write CSVs to a staging directory, BioCypher emits the
matching \texttt{neo4j-admin} headers and edge files, and the resulting
graph is loaded into Neo4j~5 via \texttt{neo4j-admin import}. The final
graph contains 6.86\,M nodes and 11.2\,M schema-typed edges across 13
node types and 34 typed relations (Table~\ref{tab:kg-stats}).

\section{Hyperparameter sweep details}
\label{app:hparam-sweep}

Hyperparameters are selected with a staged Optuna sweep that narrows
the search as model scale increases (Section~\ref{sec:setup}). Tier~1
explores broadly at the 500M scale on a single XPU; Tier~2 refines
around the Tier~1 optimum at the 3B operating point under FSDP across
four XPUs, optionally with gradient checkpointing. All trials
optimise the held-out titer probe $R^{2}$ at its best epoch, and both
studies use Optuna's \texttt{TPESampler} with median pruning.

Tables~\ref{tab:hparam-t1} and~\ref{tab:hparam-t2} list the Tier~1 and
Tier~2 search spaces in full. Trials whose \texttt{hidden\_channels} is
not divisible by \texttt{heads} are pruned at sample time.

\begin{table}[h]
\caption{Tier 1 search space (500M scale, single XPU). 1{,}000 trials,
{$\sim$}500 XPU-hours total compute budget, 80 epochs per trial,
probe evaluated every 5 epochs.}
\label{tab:hparam-t1}
\centering\small
\begin{tabular}{lll}
\toprule
Parameter & Range / set & Sampling \\
\midrule
\multicolumn{3}{l}{\textit{Architecture}} \\
\texttt{hidden\_channels}     & $\{64, 128, 192, 256\}$ & categorical \\
\texttt{num\_layers}          & $[2, 8]$                & integer \\
\texttt{heads}                & $[2, 8]$                & integer \\
\texttt{ffn\_expansion}       & $\{2, 4\}$              & categorical \\
\texttt{dropout}              & $[0.05, 0.40]$          & float \\
\midrule
\multicolumn{3}{l}{\textit{Optimisation}} \\
\texttt{lr}                   & $[10^{-4}, 5{\times}10^{-3}]$ & log-uniform \\
\texttt{weight\_decay}        & $[10^{-4}, 10^{-1}]$    & log-uniform \\
\texttt{warmup\_epochs}       & $[1, 10]$               & integer \\
\texttt{cosine\_end\_offset}  & $[1, 30]$               & integer \\
\texttt{batch\_size}          & $\{8, 16, 32, 64, 128, 256\}$ & categorical \\
\texttt{accumulate\_grad\_batches} & $\{1, 2, 4\}$      & categorical \\
\midrule
\multicolumn{3}{l}{\textit{Probe geometry}} \\
\texttt{probe\_hidden\_dim}   & $\{32, 64, 128, 256\}$  & categorical \\
\texttt{probe\_num\_layers}   & $[1, 4]$                & integer \\
\texttt{probe\_dropout}       & $[0.0, 0.3]$            & float \\
\bottomrule
\end{tabular}
\end{table}

\begin{table}[h]
\caption{Tier 2 search space (3B scale, FSDP/4). 200 trials, 30
epochs per trial, probe evaluated every 5 epochs. Initial trials
warm-started from the top-10 Tier-1 configurations.}
\label{tab:hparam-t2}
\centering\small
\begin{tabular}{lll}
\toprule
Parameter & Range / set & Sampling \\
\midrule
\multicolumn{3}{l}{\textit{Architecture}} \\
\texttt{hidden\_channels}     & $\{192, 256, 384\}$     & categorical \\
\texttt{num\_layers}          & $[8, 12]$               & integer \\
\texttt{heads}                & $[4, 8]$                & integer \\
\texttt{ffn\_expansion}       & $\{2, 4\}$              & categorical \\
\texttt{dropout}              & $[0.05, 0.30]$          & float \\
\midrule
\multicolumn{3}{l}{\textit{Optimisation}} \\
\texttt{lr}                   & $[5{\times}10^{-5}, 3{\times}10^{-3}]$ & log-uniform \\
\texttt{weight\_decay}        & $[10^{-4}, 10^{-1}]$    & log-uniform \\
\texttt{warmup\_epochs}       & $[2, 10]$               & integer \\
\texttt{cosine\_end\_offset}  & $[1, 15]$               & integer \\
\texttt{batch\_size}          & $\{32, 64, 128\}$       & categorical \\
\texttt{accumulate\_grad\_batches} & $\{1, 2, 4, 8\}$   & categorical \\
\bottomrule
\end{tabular}
\end{table}

The Tier-1 winner combines $L{=}6$, $h{=}128$ (matching the 500M row
of Table~\ref{tab:configs}) with $\mathrm{bs}{=}256$,
$\mathrm{lr}{=}2{\times}10^{-3}$, four warmup epochs, a
\texttt{cosine\_end\_offset} of~6, and all four pretraining losses
active. This configuration is used for the 500M headline run and
seeded the Tier-2 search at 3B scale.

\end{document}